# Amplifying Human Creativity and Problem Solving with AI Through Generative Collective Intelligence

**Thomas P. Kehler[1], Scott E. Page[2], Alex Pentland[3], Martin Reeves[4], John Seely Brown[5]**

[1] Chief Scientist, CrowdSmart.ai [2] Distinguished University Professor of Complexity, University of Michigan [3] Professor, Massachusetts Institute of Technology; Senior Research Fellow, Stanford Institute for Human-Centered Artificial Intelligence [4] Chairman, BCG Henderson Institute [5] Former Chief Scientist, Xerox Palo Alto Research Center**.[1*]**

**Abstract**

We propose a general framework for human-AI collaboration that amplifies the distinct capabilities of both types of intelligence. We call this, *Generative Collective Intelligence* (GCI). GCI employs AI in dual roles: as interactive agents and as technology that accumulates, organizes, and leverages knowledge. *In this second role, AI creates a cognitive bridge between human reasoning and AI models.* The AI functions as a social and cultural technology that enables groups to solve complex problems through structured collaboration that transcends traditional communication barriers. We argue that GCI can overcome limitations of purely algorithmic approaches to problem-solving and decision-making. We describe the mathematical foundations of GCI based on the law of comparative judgment and minimum regret principles and briefly illustrate its applications across domains including climate adaptation, healthcare transformation, and civic participation. By combining human creativity with AI's computational capabilities, GCI offers a promising approach to addressing complex societal challenges that neither humans nor machines can solve alone.

**Introduction**

In this paper, we propose a new framework, **Generative Collective Intelligence** (GCI), that conceptualizes AI as a social platform that facilitates the interaction of intelligent human and AI agents. The objective of GCI is to build *a dynamic cognitive bridge connecting human reasoning with AI models*.

Our framework places AI in two roles: as a participating agent in the reasoning process with unique capabilities and as a technology that accumulates, organizes, and leverages the collective knowledge produced. The second approach requires thinking about AI as a collaborative orchestrator that amplifies the distinct capacities and abilities of humans and computers. Humans possess greater awareness of human context, creativity, and grounded wisdom. Humans reason based on causal theories (Fellin Holweg 2024) based on limited data. AI, when acting as an agent, reasons based on probabilistic reasoning applied to large data. AI possesses far more impressive computational abilities and informational capacity than humans. Operating as a technology as well as an agent, AI can put those talents to use by adaptively synthesizing content as it is being produced.

---

[1*] The authors thank Allison Stanger for comments on an earlier draft of this manuscript.

Early studies of the use of AI as facilitator of human reasoning hint at the potential of GCI (Kehler et al. 2018). This has motivated our efforts in research and technology development to incorporate human cognition and advances in AI under the framing of a GCI. In what follows, we outline the importance of a cognitive bridge to building trust in AI and AI facilitated interactions. We will also describe the critical importance of adaptation, particularly in moments of rapid change.

## Why This Reframing Is Essential

In the excitement over AI's emergent capabilities, one can lose sight of the fact that AI should serves human ends. AI has been trained on human perceptions and reasoning recorded in the vast data sets used to train the foundational models. Many, therefore, view LLMs as intelligent agents. Agents like us that are often smarter than us. Given the recent emergence of human-like and better than human AI performance, that framing can be hard to avoid (Wei et al., 2022). We see it as a serious mistake that will limit progress.

The framing of AI as standalone technology that either augments or replaces human decision-making *misrepresents its nature*. AI systems are not merely independently intelligent agents, but products of human cognition embedded at every level. Generative AI relies heavily on human feedback through reinforcement learning. Newer models like DeepSeek's also depended on human cognition during early training. These systems learn from human-generated documents that embed human reasoning patterns and conceptual frameworks. Data and algorithms alone cannot address all elements of problem solving, as evidenced by the necessity of human judgment in training, prompt engineering, and problem framing. *It is critical to embrace human cognition and collective intelligence for AI to progress to the next level.*

Though AI can act like a human agent and sometimes even a super human agent, it is more productively framed as, "a new kind of cultural and social technology – a technology that allows humans to take advantage of the information other humans have accumulated" (Farrell et al., 2025). From that perspective, we see the goal of AI as producing what we call generative collective intelligence (GCI).

For GCI to succeed, AI must be able to **bridge cognition between humans and AI**. Humans must help AI and AI must help humans. We do not deny the awesome abilities of AI acting alone, nor do we deny that it improves by the minute. But it is our belief that the greatest potential of AI rests in GCI, in collaborations that combine human creativity and wisdom with AI's computational and organizational capabilities.

Our emphasis on bridging rests on the assumption that humans have something to offer, and that while AI has enhanced personal productivity in coding, graphics, and document summarization, its impact at enterprise and social levels remains modest. Achieving transformative efficiency gains and breakthroughs will require humans in the loop. With GCI, we see the potential for vast improvements in collaborations and team. We also envision more successful social groups.

## Moving Beyond Individual Productivity to Organizational Productivity

To leverage AI for the purposes of organizational and institutional development, we must include considerations of like purpose, intention, problem-framing and expectations. These aspects of problem solving processes must be considered if we are to realize the full benefits of AI to increases productivity (Reeves et al., 2024). We see coding, making graphics, and summarizing documents as but a first step.

To date, AI's impact at the business enterprise level or social level has been modest. To achieve large increases in efficiency and scientific breakthroughs requires a ***reframing that places human interactions within collaborations, teams, and social groups at the core.*** Those interactions are currently not prominent in current thinking about AI. We believe they are critical to the next generation of AI.

*We also believe that **humans must specify human ends and that humans must collectively align on achieving those ends.*** AI cannot find an answer if it does not know the question. In many instances, answers to important questions are not always present in pre-existing knowledge. Comprehension of current information and its implications to achieving human goals in the context of a specific situation require human judgment and experience. And, not just from a single human. Solutions will require diverse perspectives and ways of thinking to produce collective intelligence.

## A vision and agenda for GCI

GCI offers far greater potential than human-only collective intelligence. Human collective intelligence takes two forms. It can be *super additive.* Collections of talented, diverse people can outperform any one of them acting alone. It can also be *emergent.* Collections or groups can perform tasks or functions that exceed the capabilities of any one person. No one person can perform a heart surgery, operate a nuclear submarine, or perform as a jazz ensemble.

Society leverages human collective intelligence to drive progress. We use groups to make important predictions and decisions and teams to do scientific research. The scientific method of knowledge discovery is a deliberative process involving collective intelligence. Collective intelligence operates at scale in the (mostly) accurate prices that emerge in financial markets. Aligning on evidence is the result of a deliberative process of exploring areas of friction and disagreement. *GCI accommodates the resolution process of balancing divergent and convergent thinking*.

Collective intelligence requires diverse, capable people interacting in structures appropriate to the task. Collectively intelligent problem solving requires deep, probing interactions. We therefore claim that the greatest potential of AI lies not in its capacity to act alone but in its ability to bridge human creativity and wisdom with AI's computational and organizational capabilities. This reframing acknowledges AI as "a new kind of cultural and social technology" that allows humans to leverage collectively accumulated information while maintaining human cognition at its core.

This new framing of AI rather than extractive and disconnected from the dynamics of individual and social learning combines the capabilities of humans and AI. AI's capacity to instantaneously organize enormous amounts of data, information, and knowledge in multiple formats allows for entirely new modes of interaction and has immense untapped potential for a new wave of AI.

Frameworks that expand perspectives have special promise in the modern, whitewater world where change and complexity rule. These new frameworks can give information seeking and adaptation precedence over the codification and ritualization of processes and procedures that predominate in human run organizations.

We envision GCI as an adaptive, learning supervisor that incorporates both human and AI reasoning. Humans and AI do not "think" alike. The architecture we describe leverages the diverse abilities of AI and humans. AI can process huge amounts of information. Humans can provide logical checks as well as apply creativity and wisdom. The limitations of recall, a known problem in humans, is more than compensated for in an LLM's ability to present "the world of human knowledge" at our fingertips. The greater power is in leveraging this complementarity.

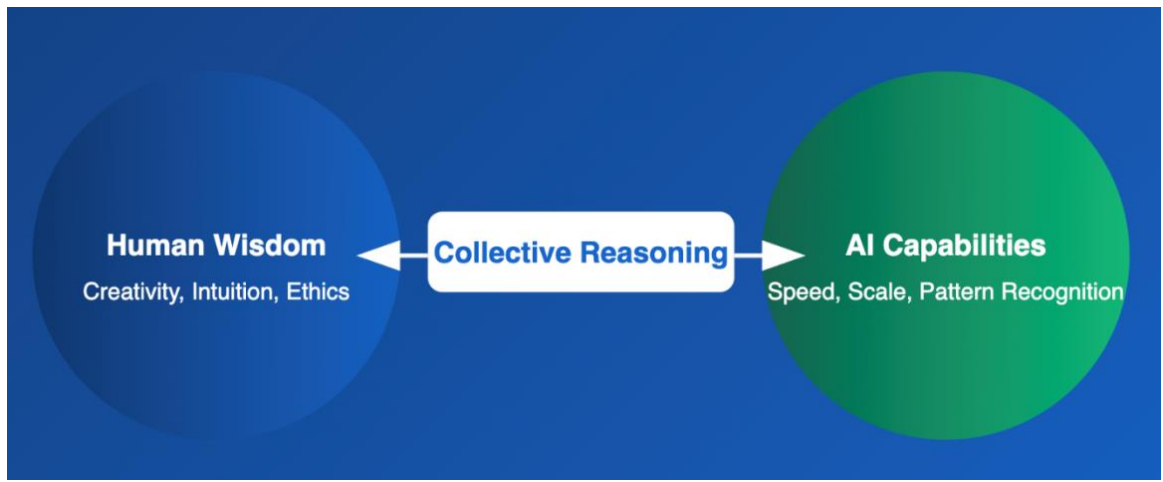

Collective reasoning provides a powerful framework for people and AI technology to collaborate and deliberate together *enabling disparate groups of any size to share ideas and make informed decisions*. By combining human creativity and wisdom with AI's ability to organize information allows us to overcome communication barriers, find common ground, and develop solutions to complex problems that neither humans nor AI can solve alone.

GCI and collective reasoning assistants can amplify human ingenuity, foster serendipitous discovery, and catalyze epiphanal breakthroughs in addressing our most pressing challenges. In it, we explore use cases that demonstrate the intrinsic value of GCI. Collective reasoning provides a cognitive bridge between generative AI and human problem solving.

For AI to evolve to the next level it is essential that we create a robust framework that enables people and technology to collaborate. ***Instead of assuming the current LLMs can reproduce our creativity, logic and ingenuity, the GCI framework extends generative AI by it as a computationally gifted agent with enormous capacity working in conjunction with human agents who possess greater social awareness, creativity and reasoning.*** This is a co-creative partnership partially recognized by generative AI in filtering the generative outputs of models. Human judgment is used in reinforcement learning to make the results at least plausibly intelligent. GCI makes the need for human cognition explicit and embraces it in the design of the architecture radically expanding the potential benefits.

## The Foundation of Generative Collective Intelligence

GCI takes as its starting point, AI pioneer Judea Pearl's observation that: "You are smarter than your data. Data do not understand causes and effects; humans do." We add to that observation the idea that much of intelligence is social or collective. The scientific revolution was and is a social phenomenon. Thus, we must frame AI from the perspective of the collective for it to flourish and be a trusted part of our continued evolution.

Trust, built on aligned objectives, is key. While generative AI has made remarkable strides in processing vast amounts of information, its opaqueness does not always engender trust. It is weak on contextual understanding without extensive prompting, and lacks the creative intuition that humans possess. Current AI systems can suggest possibilities, but these suggestions cannot easily lead to action because they lack trustworthiness—a quality that emerges from transparent collaboration, source verification, and collective validation. That trustworthiness stems in part from human's inability to correctly infer the confidence of AI's statements. AI presents things that might be true in the same language as things that are almost always true. Such features create a frustrating, suboptimal learning environment (Lehmann et al., 2024).

The GCI approach we describe is grounded in fundamental principles. The first building block of GCI is the Law of Comparative Judgment, developed by psychologist Louis Leon Thurstone in 1927. This law, which is more accurately described as a model or method, scales subjective attributes based on comparing pairs of stimuli. Those comparative judgments produce scores for each stimulus rather than measurements of actual physical properties. If you wear eyeglasses, it is the repeated application of this test that is translated into a prescription.

The comparative framework mirrors human cognition —we understand concepts not in isolation but in relation to other concepts, making learning more intuitive and effective as students build mental models through structured comparison rather than rote memorization of disconnected facts.

For a more detailed example, imagine we have ten people compare three shapes for a spatula handle. Call these A, B, and C. The method has person compare A to B, A to C, and B to C. If seven people preferred A to B, six preferred A to C, and eight preferred B to C, the pairwise rankings include inconsistencies. People are human, so that will almost always be the case.

Thurstone's method uses maximum likelihood to infer values for each of the alternatives under the assumption that the probability of choosing A over B is proportional to the value of A over the value of B.[2] In our example, Thurstone's method would return values of approximately 0.5, 0.35, and 0.15 for A, B, and C respectively.

It should be clear why the Law of Comparative Judgement is so important. It moves us from ordinal rankings to cardinal scores. As new information and rankings arise, those scores can be adjusted.

The second building block of GCI is the ability to generate ideas and to decide wither to test them. Humans, and indeed all animals, confront an exploitation-exploration dilemma (Hills et al 2015). They must decide whether to stick with familiar ideas or whether to expend effort finding new, better ideas (Kahneman, 2011). Humans, and all social animals, have been found to engage in social learning in these contexts. They look to others for ideas.

---

[2] Note that formally, this is the Bradley-Terry Method. The example is illustrative. The dynamic interpretation of the law of comparative judgment uses Bayesian mechanics and a Hidden Markov Model.

This necessity to combine personal and community evidence brings us to the principle of minimum regret as a method for selecting between alternatives. This principle states that an agent or group of agents makes decision at each instant that minimizes the worst case regret.

Recent work by cognitive and AI scientists indicate that the minimum regret principle can explain how some groups of humans learn from each other and choose what actions to take (Krafft et al., 2021; Pentland, 2013; Adjodah et al., 2021).

Minimum regret decision making can be applied to multi-armed bandit problem in which an individual chooses one of a discrete set of alternatives in each period. In the last decade, mathematical solutions to such problems have been extended to distributed agents; for example, a gambler observes the payouts of other casino patrons and combines those observations with personal knowledge to decide which slot machine to try next. This is a clear analog to individuals in a group observing others to learn what does and does not work in that environment. Such models can also include competition, where agents split the payoff if both choose the same arm (Liu and Zhao 2010).

The minimum regret mathematical results offer a strategy that a group of agents may use to form an optimal policy for action. The core of this strategy is for each person to use their community to provide new alternatives for personal consideration. They can do so using the Law of Comparative Judgment to select alternative.

In seeking to minimize regret, people experiment with alternatives that have uncertain rewards and even try actions that are unfamiliar. It is this necessity to explore for new ideas that gives rise to the exploit-explore dilemma.

In animals, the operation of minimum regret in driving exploration is known as foraging. Animals have routines for finding food and water, but sometimes they explore in the hopes of finding yet better sources for food and water. When circumstances are uncertain, animals respond by increased exploration. While exploration is wasteful in terms of immediate access to resources, it is key for minimizing regret…. you never know when you will find something better.

For a more concrete application, consider the example of bees searching for flowers. At the start, a hive member makes the decision to leave the hive and search for a new flower bed (i.e., tries an unfamiliar and uncertain action). Most other bees observe the excitement of that bee when it returns (so they can estimate the likelihood of this being a significant discovery) but have also learned through experience that great new flower beds are rare (i.e., the prior probability is low) and decide not to try out the new flower bed (i.e., their posterior probability used to select actions is to do nothing). However, some bees do try the new flower bed (perhaps young ones with less experience and more optimistic prior probabilities), and as more and more returning bees are seen to be excited, the likelihood eventually overwhelms the prior, and most of the bees begin to visit the new flower bed.

This strategy for optimal, minimum regret decision-making has demonstrated excellent performance in many domains, enough so that it is a standard approach in signal processing, medical decision-making, and finance. It generalizes well to to new and changing situations and the ability to work with noisy, long tailed, and ill-conditioned data inputs (Dubey and Pentland, 2020a). If the agent is smart enough to be capable of a little mental reflection, it can also handle situations where some agents have different, even adversarial, goals. Agents decide who they should trust to be part of their learning (story-sharing) network by comparing the choices they have made to the choices of the other agents and, from that comparison, select the best subset of agents with whom to trade stories (Dubey and Pentland, 2020b).

## Implications

The main implication of our framing *GCI embraces and extends the capabilities of generative AI.* Current AI models would have substantially greater problems with hallucination without the post processing alignment systems that rely on human judgment. In addition, DeepSeek gained radical improvements on efficiency of learning by applying human judgment to the training process. While highly useful, neither of these approaches tap into the 'why' behind why a particular generative output of a model is 'not plausible or acceptable'. Proper application of the comparative judgment and minimum regret offers a treasure trove of innovations in the building of foundational models.

The reframing of AI as a bridging technology and as an agent has profound significance across education, deliberative democracy, and collective intelligence. For AI systems and collective intelligence applications, the Law of Comparative Judgment and Minimum Regret provide mathematical foundations for preference learning, enabling machines to understand human values and priorities in nuanced ways. Together, they establish a robust framework for aggregating diverse human judgments into coherent knowledge models, regardless of group size or problem complexity. Embracing established models of cognition is mission critical to the quality, trustworthiness and utility in critical decision making in future AI systems.

The framework also handles uncertainty by helping AI systems represent confidence levels in human decision-making processes, acknowledging that not all judgments carry equal certainty. Perhaps most practically, it leverages the sophisticated A/B testing methodologies that underpin recommendation systems, prediction markets, and decision support tools that have become ubiquitous in our digital landscape. When implemented within collective intelligence systems, this approach transforms subjective human wisdom into structured insights that are incorporated into persistent knowledge models, preserving their essential human qualities.

A critical difference in GCI is the use of a dynamic comparative judgment and minimum regret approach to quantify values and risks associated with actions. This allows human groups find common ground and create intentional futures.[3] GCI, thus, provides a scalable way to catalyze how individuals and groups resolve uncertainty in decision-making, planning, and collective discovery of new paths forward. The GCI framework embraces the complementarity in human and machine intelligence.

We have barely scratched the surface of understanding the power of the human brain. Currently, it requires massive amounts of energy, time, and money to develop a foundational model for GenAI. The human brain operates on roughly 15 watts, and its potential is vastly underutilized in current AI systems. By coupling human creativity with the extreme power of GenAI, GCI unlocks vast new possibilities for a new generation of AI applications that promise a revolutionary impact on both social and commercial fronts. Keep in mind that GCI acquires and retains generative models of the knowledge generated by this coupling of the power of human creativity with the power of GenAI.

---

[3]. The human brain relies on similar processes to organize information and generate action.

In education, GCI can transform assessment and learning by enabling evaluation methods that capture nuanced understanding beyond simplistic binary right/wrong paradigms. It can illuminate conceptual relationships between ideas as students engage in structured comparison of alternatives, naturally developing the critical thinking skills essential for complex problem-solving. Through the patterns of preferences revealed in comparative judgments, educators can gain insight into students' cognitive processes, allowing for the creation of personalized learning paths tailored to individual conceptual development.

GCI can also provide a scientific foundation for preference aggregation to facilitate deliberative democracy and increase civic engagement. . Rather than just counting votes, GCI allows for measuring the intensity and consistency of preferences across a population. GCI enables a more nuanced understanding of opinion and perspectives where the strength of preferences matters, not just their direction.

In political contexts, disagreement often stem from differences in instrumental preferences. People often share end goals - less poverty and crime, a cleaner environment, greater economic growth.  GCI can facilitate deeper reasoning to create shared understanding of how policies might work. A voter may prefer a new highway based on linear reasoning that it will result in less traffic. That voter can be presented with system level logic and evidence that challenges that assumption.

In theory, GCI's use of a dynamic comparative judgment and minimum regret can transform subjective human judgments into objective measurements incorporated in a computational form that preserves the richness of human intuition. This bridge between qualitative human judgment and quantitative machine processing is essential for creating AI systems that genuinely augment human intelligence rather than merely mimicking or replacing it.

GCI can integrate research in collective intelligence which specifies how different models and perspectives can be combined to create greater predictive accuracy. GCI guides, captures and builds a map of relationships based on the context of the reasoning. By mapping verbal descriptions into knowledge maps, GCI can distinguish model diversity from predictive diversity. Diverse models can be tested and improved. Diverse predictions from the same model can be adjudicated. Often, they result from wrong "facts."

### Amplifying Serendipitous Discovery

Finally, GCI can amplify serendipitous discovery—unexpected connections and insights that emerge when diverse perspectives interact in structured ways (Fink et al., 2017). Traditional group decision-making often falls victim to combinatorial complexity, as the increased number of participants creates exponentially more possible subgroups. From among a group of twenty people, which group of three would be most likely to produce a breakthrough?  There are more than a thousand possible groups.

A Collective Reasoning Agent can address this challenge by using the minimum regret principle to orchestrate collaboration so that it highlights disparate perspectives that have sufficient common understandings to communicate with one another.  That agent can also mask identities and scores during deliberation so that participants engage directly with ideas rather than status or preconceptions, creating an environment where the most beneficial connections can emerge without social frictions.

The masking of identities during deliberation creates what psychologists call "psychological safety"—an environment where participants feel free to contribute unconventional ideas without fear of judgment. This psychological safety, combined with structured facilitation, creates ideal conditions for serendipitous discovery. Additionally, masked identities transcend traditional hierarchies, prioritizing the power of the idea over the power of one's identity or position.

In practice, we expect environmental safety to accumulate over time as people learn that the architecture emphasizes ideas not credit assignment or blame. Sharing creative ideas presents no possibility of ridicule. The idea just gets voted down.

In addition to masking identities, participants are unable to effectively promote their ideas because ideas are not weighted by status hierarchies. The system adaptively selects items for peer review. For this reason, it is nearly impossible to game the system, thereby protecting against manipulation and groupthink.

## The Power of Different Thinking Styles and Perspectives (Cognitive Diversity)

One of the key findings of the science of collective intelligence is that cognitively diverse groups of subject matter experts with different backgrounds and perspectives tend to be more accurate in their predictions than any single expert. When harnessed adequately through a Collective Reasoning Agent, diverse perspectives become a powerful asset rather than a communication challenge.

Consider a civic planning challenge, such as redesigning urban transportation systems for sustainability. Within a GCI framework, transportation experts, environmental scientists, community advocates, business leaders, and technologists would contribute their perspectives on criteria like accessibility, environmental impact, economic feasibility, and community needs. A Collective Reasoning Agent can facilitate this exchange, helping synthesize insights across disciplinary boundaries while identifying areas of alignment and divergence. This is a particular area where AI agents could become aids in the planning process. This type of framework is used in several cities to find better consensus plans (MIT City Science).

In this process, a transportation engineer's technical solution might unexpectedly resonate with a community advocate's accessibility concern. At the same time, an environmental scientist's caution might inspire a business leader to propose an innovative funding model. These serendipitous connections—unlikely to emerge in traditional, siloed planning processes—become visible and actionable through GCI.

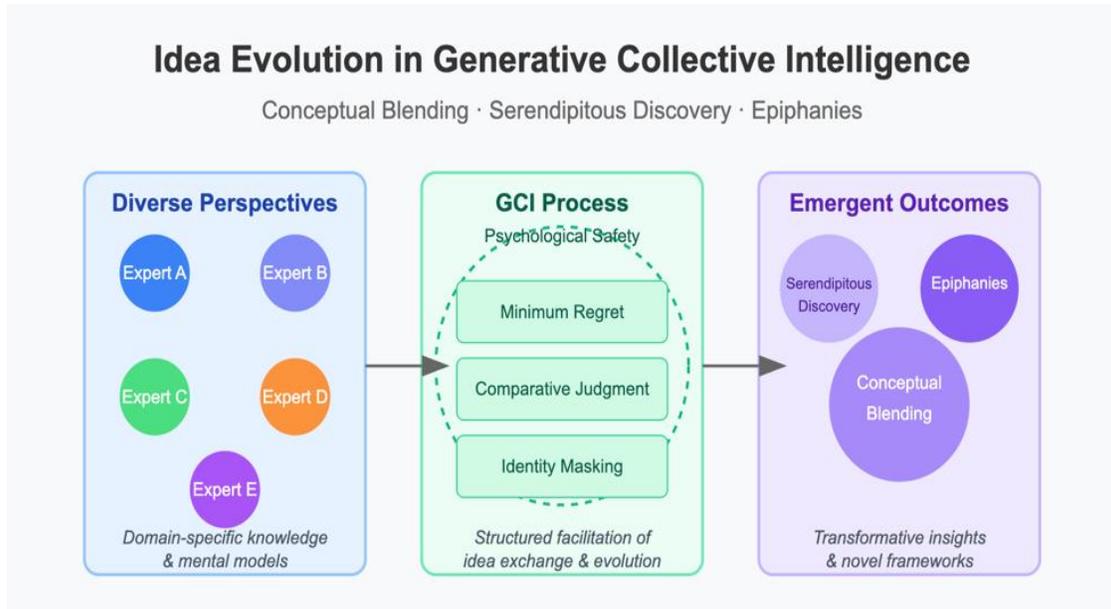

Figure 2: Idea Evolution in Generative Collective Intelligence

## Catalyzing Epiphanal Breakthroughs

*From Incremental to Transformative Solutions*

Beyond serendipitous discovery lies the potential for true epiphanies—transformative insights that fundamentally reframe our understanding of problems and solutions. These transcendent breakthroughs often emerge at the intersection of seemingly unrelated domains when different points of view and points of constructive friction resolve through collective reasoning. It is here that the power of the complementarity of human intelligence and AI agents comes into full fruition. An LLM's recall and generative capabilities link to collective human experience and imagination to enable profound possibilities. The human mind excels at imagination and there remains immense potential in amplifying this capability (Reeves and Fuller, 2021).

A Collective Reasoning Agent can facilitate these breakthrough moments by identifying and highlighting tensions between different perspectives. Rather than seeking premature consensus, it guides participants through a process of exploring these tensions, creating the cognitive conditions necessary for epiphanies to emerge.

In addressing climate adaptation challenges, for example, a breakthrough might emerge when indigenous knowledge systems about ecological balance are brought into dialogue with advanced climate modeling techniques. The resulting insight—a hybrid approach that combines data-driven forecasting with generations of observational wisdom—represents more than the sum of its parts. It's a fundamentally new framing that neither approach could achieve independently.

## The Architecture of Epiphany

GCI's approach to breakthrough thinking seeks to create what cognitive scientists call "conceptual blending"—the process of integrating elements from different mental spaces to create new conceptual structures. A Collective Reasoning Agent can facilitate this blending by:

1. Identifying resonant themes across diverse perspectives
2. Making connection patterns visible that might otherwise remain hidden
3. Creating space for iterative refinement of emerging insights
4. Seeking to challenge individual bias and avoid groupthink
5. Building causal models that represent shared understanding

This process allows participants to transcend their initial frameworks and arrive at genuinely novel solutions. The resulting breakthroughs are both innovative and practical because they emerge from collective deliberation rather than isolated ideation.

## Application to Societal and Civic Challenges

Generative collective intelligence offers transformative potential across numerous societal and civic challenges that demand both innovative thinking and broad stakeholder buy-in:

### Climate Adaptation and Resilience

Climate change requires both technological innovation and social transformation. A GCI approach brings together climate scientists, policy experts, industry leaders, community organizers, and citizens to develop adaptation strategies that are scientifically sound, economically viable, and socially acceptable. A Collective Reasoning Agent can help identify shared priorities while surfacing innovative approaches that might otherwise remain buried in siloed conversations.

### Healthcare Access and Quality

Healthcare systems worldwide face challenges of cost, access, quality, and equity. GCI facilitates collaboration between medical professionals, patients, insurers, public health experts, and technology developers to reimagine healthcare delivery. By capturing the collective intelligence of diverse stakeholders, the system can identify breakthrough approaches that balance competing priorities while meeting shared goals.

### Educational Transformation

Education systems must evolve to prepare learners for rapidly changing futures. GCI enables educators, students, parents, employers, and policymakers to collectively reimagine educational models. The serendipitous discovery process might reveal unexpected connections between workplace needs, pedagogical approaches, and student engagement strategies, leading to transformative educational models.

### Civic Participation and Governance

Democratic systems benefit from broad and meaningful civic participation. GCI can facilitate inclusive deliberation at scales previously unimaginable, allowing thousands of citizens to contribute to policy development in structured ways. A Collective Reasoning Agent can help identify areas of consensus, clarify trade-offs, and surface innovative governance approaches that emerge from collective wisdom.

## Exploring Breakthrough Solutions with GCI

Earlier we presented a preliminary example of how GCI works as an episodic experience. The power of this framework is best seen in using as a continuous way to discover, imagine and predict possible solutions through an abductive reasoning process. Like Sherlock Holmes we can consider the probabilities of which of several possible candidate solutions might work.

Part of an advanced GCI architecture is the ability to guide a group of human and AI agents through a scripted process enabling the group to assign probabilities of success to different solutions. Specifically, the group is guided through a multi-criteria decision model for different solution candidates each evaluated according to a set of criteria.

Consider the healthcare example above. Healthcare is a global problem. It is a topic discussed for the past several decades. An initiative led by the 90_10 Institute seeks to create a 'De Novo Design of a New Biotech for Society'. To do this they are engaging the collective and collaborative participation of each of the elements of the ecosystem for Biotech V2.0 in an early application of GCI. The participants include biotech, pharma and diagnostics companies. It also includes health impact motivated funders, venture capitalists, insurers, innovators in private equity, social impact bonds, government treasury, payers, patient advocacy groups, public health institutions and governments as policy promoters and payers.

The engagement is operating under a privacy policy with the participants. For that reason, we cannot show details of the engagement. GCI was used to engage the various groups. One result the system produces is a prioritized list of the 'collective voice' of the group ranked by the probability of shared importance. Video conference calls were used to allow the group to review the interim results. Quite often, because of reviewing the collective voice, participants would reengage to add new ideas or build on other emerging themes. The group's work is captured in a private language model of their deliberations which can be queried using the conversation capabilities of generative AI. Unlike large models, the only information in the model is from the deliberations and it is fully provenanced and trustworthy. During the deliberation a root cause analysis led to a possible 'blue ocean' idea going forward.

The diagnosis was that 'the problem with the system is that it treats healthcare as an expense. What if we treated healthcare as an asset class instead?' This is an example of a potential breakthrough idea. The difference with GCI we move from creative idea to innovative development with a group of experts is that help resolve those issues that often cause great ideas to die on the vine.

> **Prompt: What steps would need to be taken to consider the patient as an asset class, shifting the entire economic model?**
>
> *To consider the patient as an asset class and shift the entire economic model, several steps would need to be taken. Firstly, there needs to be a fundamental shift in mindset within the finance community to stop viewing healthcare solely as a cost and start recognizing it as an asset. This shift would involve creating a derivatives market in healthcare, similar to other sectors, allowing for the transformation of healthcare products into tradable assets. By establishing a derivatives market, healthcare entities could hedge against future uncertainties, thereby changing the perception of healthcare from a cost center to an asset class.*
>
> *Additionally, there would need to be a focus on developing innovative financial instruments specific to healthcare, such as health derivatives, that would enable stakeholders to manage risks effectively. This would involve educating both financial professionals and healthcare executives on the benefits and mechanisms of these instruments to facilitate their adoption.*

While the deliberation is done in a 'single blind' mode, the identity of contributors is maintained allowing engaging with individuals or subgroups on emergent ideas. In every deliberative conversation the system creates a prioritized list of who contributed the most relevant knowledge based on peer reviewed results. This allows subgroups to form around topics. *GCI supports a new class of AI empowered learning communities.*

From here the group can form proposed initiatives, learn the groups confidence in their success, debug problems and re forecast alternatives. GCI supports the development of scalable learning communities that have significant untapped potential.

## The Path Forward: Building Trust Through Generative Collective Intelligence

For generative collective intelligence to fulfill its potential in addressing societal challenges, trust remains the essential element. The GCI approach builds trust through:

1. Transparency of process: A Collective Reasoning Agent can create a comprehensive audit trail of how knowledge models are constructed.
2. Attribution of ideas: The system maintains provenance of information, ensuring that *contributions remain traceable*.
3. Explanation of outcomes: The causal knowledge model provides explanations for recommendations.
4. Iterative improvement: Continuous learning from outcomes creates a feedback loop for better decision-making.

By combining the creativity and judgment of humans with the organizational and analytical capabilities of AI, generative collective intelligence offers a path toward addressing our most complex societal challenges. It provides a framework where serendipitous discovery and breakthroughs can emerge through structured collaboration, creating solutions that are both innovative and trusted.

As we face unprecedented global challenges, this integration of human wisdom and technological capability may be our most promising path forward—not by replacing human judgment with artificial intelligence, but by creating a symbiotic relationship where each amplifies the other's strengths while compensating for limitations.

## Conclusion

GCI represents more than a technological advancement; it offers a new paradigm for collaborative problem-solving that honors human creativity while leveraging AI capabilities. Through structured facilitation by a Collective Reasoning Agent, diverse stakeholders can overcome communication barriers, discover unexpected connections, and achieve breakthrough insights that transform how we approach our most pressing societal challenges.

The power of this approach lies in its ability to create the conditions for both serendipitous discovery and epiphanal breakthroughs while building the trust necessary for collective action. As we navigate an increasingly complex world, generative collective intelligence provides a framework for harnessing our collective wisdom to create the future we aspire to build.

## A call to action

A common narrative on AI suggests that with more data, more GPUs, more money, we are on the road to AGI.  It is time for that narrative to change. The next wave of AI taken from a social and cultural frame needs support at all levels including awareness, education, research, commercial implementations, and social experiments in a new type of engagement that is constructive and aims to promote the common good. The additional benefit is the energy and cost efficiency adaptive learning technology – it is more in line with the energy efficiency of the human brain.

## GCI is available for deployment

There are a growing number of deliberation platforms such as Polis, Frankly, and deliberation.io that have been designed to promote collaboration, common understanding and common ground. In most of these platforms, AI facilitates but does not operate as an agent as well.   These programs exist at various levels of readiness.

We believe that alignment on priorities for the  tasks associated with innovation and deployment will require a next level of GCI has the capacity for scenario forecasting and iterative design improvement.  We see a logical disconnect between the claims that AI can help us to solve hard problems and then not using AI to help design those processes, which is itself a hard problem.

CrowdSmart is a pioneer in using GCI for decision intelligence and prediction. The GCI examples discussed in this paper are based on CrowdSmart's technology. Early on, CrowdSmart pioneered GCI in an investment context, predicting survivability of startups. Following that early experiment, a system was released that addressed a broader set of use cases. CrowdSmart provides and continues to develop a commercial platform that enables the application of Generative Collective Intelligence to a wide range of commercial and social impact applications. CrowdSmart is transitioning to an open-source code and open API to work with a broader AI development community in business applications, government applications, and academic research.

Common Good AI is a nonprofit organization with the mission to foster inclusive civic engagement by transforming how communities find common ground and solve problems together using new digital tools. They enable the application of new AI technologies and tools to find common ground for the common good.